\documentclass{llncs}
\usepackage{mathptmx}       % selects Times Roman as basic font
\usepackage{helvet}         % selects Helvetica as sans-serif font
\usepackage{courier}        % selects Courier as typewriter font
\usepackage{type1cm}        % activate if the above 3 fonts are not available on your system
\usepackage{makeidx}         % allows index generation
\usepackage{graphicx}        % standard LaTeX graphics tool when including figure files
\usepackage{multicol}        % used for the two-column index
\usepackage[bottom]{footmisc}% places footnotes at page bottom
\usepackage{subfigure}
\usepackage{amsfonts}
\usepackage[cmex10]{amsmath}

\begin{document}

\title{Let's talk about love: use of explicit replies as coordination mechanisms in online student debates}
\titlerunning{Let's talk about love}  % abbreviated title (for running head)
%                                     also used for the TOC unless
%                                     \toctitle is used
%
\author{Bruno D. Ferreira-Saraiva\inst{1} \and João P. Matos-Carvalho\inst{1,2} \and
Manuel Pita\inst{2,1} }
\authorrunning{Ferreira-Saraiva et al.} % abbreviated author list (for running head)

\institute{COPELABS, Universidade Lusófona \\
\and
CICANT, Universidade Lusófona \\
\email{manuel.pita@ulusofona.pt} \\
Campo Grande 388, 1700-024.\\
Lisbon, Portugal}

\maketitle    

\section*{Introduction}
%
%Recent research has approached conversation from the lens of dynamical systems theory \cite{fusaroli_2014,fusaroli_2016}. 
%
People in conversation entrain their linguistic behaviours through spontaneous alignment mechanisms \cite{Pickering:2004aa}---both in face-to-face and computer-mediated communication (CMC) \cite{Scissors:2009aa}.
In CMC, one of the mechanisms through which linguistic entrainment happens is through \textit{explicit replies}.
%
%For example, people on Facebook or WhatsApp often use explicit replies to agree or oppose other people and their ideas.
%
%Explicit replies can also provide support, attack or elaborate on what others have said.
%
Indeed, the use of explicit replies influences the structure of conversations, favouring the formation of reply-trees typically delineated by topic shifts \cite{okamoto_2001}. 
The interpersonal coordination mechanisms realized by how actors address each other have been studied using a probabilistic framework proposed by David Gibson \cite{gibson_2003,gibson_2005}.
Other recent approaches use computational methods and information theory to quantify changes in text.

We explore coordination mechanisms concerned with some of the roles utterances play in dialogues--specifically in explicit replies.
We identify these roles by finding community structure in the conversation's vocabulary using a non-parametric, hierarchical topic model.
Some conversations may always stay on the ground, remaining at the level of general introductory chatter. 
Some others may develop a specific sub-topic in significant depth and detail. 
Even others may jump between general chatter, out-of-topic remarks and people agreeing or disagreeing without further elaboration. 
%
%The study of mediated online conversation \cite{Herring:1996aa} is relevant in our current hyper-connected social world since it influences, e.g., what people choose to believe or their political inclinations, and interpersonal coordination in these scenarios is different from regular face-to-face interactions \cite{Meredith:2020aa}.

Popularized by platforms like Twitter and WhatsApp, short-text communication has arguably led to a decline in profound, meaningful conversations among youth. 
For example, Vaterlaus \emph{et al.} noted short-text interaction not only limits the depth of discussion but also undermines the development of critical thinking skills \cite{Vaterlaus:2016aa}. 
%
%Furthermore, the culture of instant gratification increases impatience and intolerance towards detailed, nuanced discussions \cite{Carr:2020aa}. 
%
In addition, the like-dislike functionality on social media platforms fosters polarization, and the formation of echo chambers, wherein users are primarily exposed to viewpoints aligning with theirs \cite{Pariser:2011aa}. 
The interactions favoured by social platforms can lead to a fragmented society, with individuals less likely to engage in fruitful dialogues with those holding opposing views, promoting intolerance and misunderstanding \cite{Sunstein:2018aa}. 
%
%Robin Alexander defends that dialogical teaching is critical to responding to the current cultural crisis, but only if it considers the most strongly attacked elements: language, voice, argument, and truth \cite{Alexander:2019aa}. 
%
We ask (a) To what extent do high-school students use explicit replies in text-based, multi-party, task-oriented group debates? (b) What are the functions of explicit replies? (c) Do different groups use replies in consistently different ways? If so, what roles may explicit replies play in collective information processing, entrainment and synergy?

We collected data from 25 multi-party student debates on the question ``Is being in love like a disease?'' in an online educational setting. 
We used purposefully designed chat rooms where (a) participants were pseudonymous, as they used randomly generated user names despite belonging to the same school groups; (b) interactions were text-based; (c) the interface had the explicit reply affordance. 
The data consisted of chronologically ordered sequences of text turns. 
Each turn is a node, where each reply node is linked to its parent node.
We obtained a topic model for the entire corpus, which comprises all turns in the 25 debates.
The chosen topic-modelling algorithm \cite{Gerlach:2018aa} represents the input corpus as a bipartite network. 
In this network, one class of nodes represents turns, while another represents words from the corpus vocabulary. 
The algorithm identifies the best modular partition of the network by using stochastic block modelling \cite{Karrer:2011aa}, iteratively merging modules to produce a hierarchical topic model. 
This topic modelling algorithm has three key advantages: (a) it does not require the pre-specification of the number of topics; (b) it does not assume that topics have idealized distributions of word frequencies; (c) it was validated on short-text corpora.

\begin{figure}
\centering
\includegraphics[width=4in]{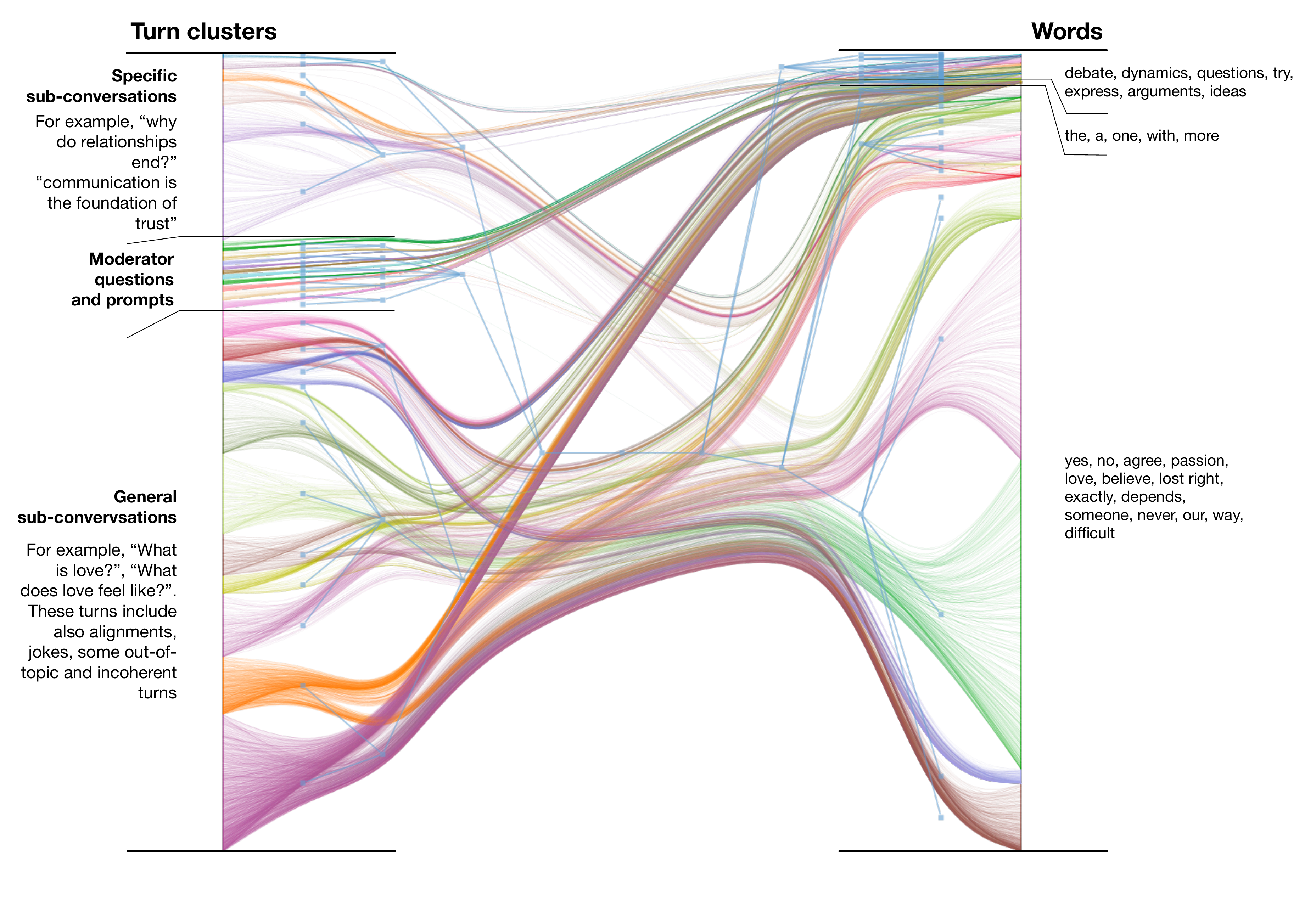}
\caption{The topic model obtained for the 25 debates on love as disease.}
\label{fig:topic_model}       % Give a unique label
\vspace{-0.4cm}
\end{figure}

The topic model assigned input turns to topic-mixture clusters (see Figure \ref{fig:topic_model}). 
We used qualitative content analysis to label document clusters according to whether they corresponded to (G) general talk about love as disease, (S) specific sub-topics within the general subject, (A) alignment expressions, (I) incoherent turns, including out-of-topic and text written in other languages; (MP) moderator prompts, (MQ) moderator questions, and (U) turns that were not considered for the topic model, e.g., because they were too short.
In the last data pre-processing step, we validated label assignment by leveraging the chatGPT API with a purposely designed prompt.
With the labelled data, we then computed the empirical probability distribution of explicit replies over all labels for each debate. Next, we calculated the similarity matrix for the debate rooms using a distance formulation of the Jensen-Shannon (JS) divergence between two probability distributions, $P$ and $Q$, given by the equation,

\begin{equation}
\label{Eq:CumulSpecFnc}
JS_{\text{distance}}(P, Q) = 1 -\sqrt{\frac{1}{2} \sum_i P(i) \log_2 \left(\frac{P(i)}{M(i)}\right) + \frac{1}{2} \sum_i Q(i) \log_2 \left(\frac{Q(i)}{M(i)}\right)}
\end{equation}

\noindent
$\text{where } M(i) = \frac{1}{2} \left(P(i) + Q(i)\right)$. 
Finally, we clustered the debate rooms using hierarchical clustering, setting the cut-off value to $JS_\text{distance} = 0.5$.

\section*{Results}

\begin{figure}
\centering
\includegraphics[scale=0.65]{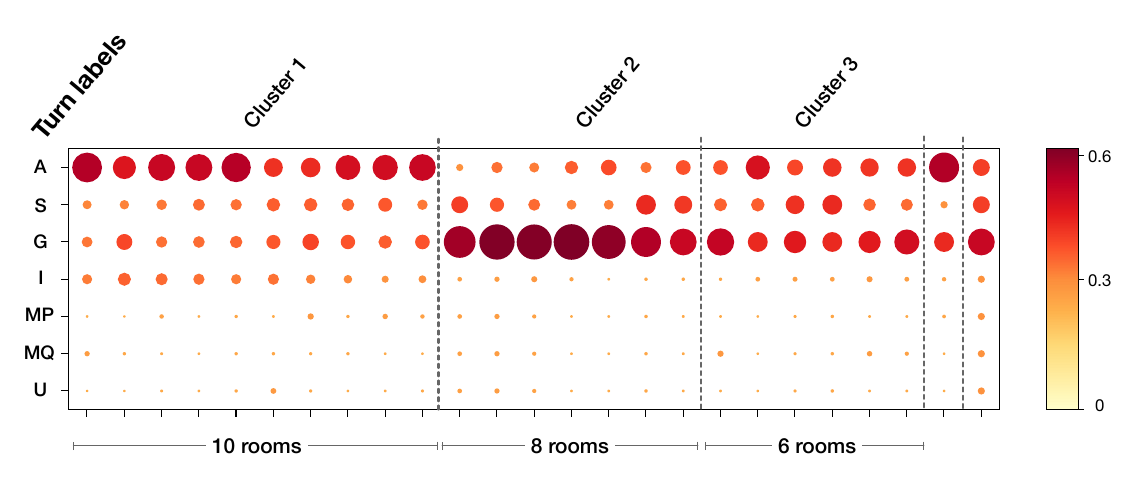}
\caption{The empirical probability distributions of replies over labels, partitioned in three clusters (two debates were not assigned to any cluster)}
\label{fig:clusters}       % Give a unique label
\vspace{-0.4cm}
\end{figure}

356 students participated in the debates, where participation per debate ranged from (min) 6 to (max) 27 students, and 14 on average. 
Across the 25 debates, reply turns accounted for 45.6\% of the utterances, with a standard deviation of 12.4\%. 
We found no significant differences in the number of explicit replies that we could attribute to the number of turns or debate participants.
After computing the empirical probability distributions of reply counts, we partitioned them into three clusters, see Figure \ref{fig:clusters}. 
The first cluster consisted of ten debates where students used replies primarily for alignment, e.g., expressions such as ``I agree.'' or ``Exactly!''. 
The second cluster included eight debates where explicit replies were mostly generic chatter about love as a disease.
Students in these rooms used replies to create conversational continuity in ways similar to the first cluster, but where the alignment is realized through paraphrasing or elaborating on previous utterances. 
The third cluster grouped debate rooms in which alignment and generic chatter were the main functions for which explicit students used explicit replies.

Even though students rarely used replies for incoherent (out-of-topic) utterances in these debates on the subject of love as a disease, this specific use of replies was twice as high in the in the first (alignment) cluster.
Indeed the overall proportion of incoherent remarks suggests students were engaged with the subject.
Similarly, we noted that the use of replies for deepening the conversation was slightly higher in clusters 2 (generic chatter) and 3 (mixed alignment and generic chatter).

Remarkably, the variable participation levels in these debates, ranging from 6 to 27 students, did not significantly affect the use of explicit replies, emphasizing the consistent engagement levels across different scenarios. 
The interpersonal coordination implemented through the use of replies was focused on social ties.
It is intriguing that the alignment-centric debates, characterized by consensus and agreement, also bore the highest instances of incoherent replies. 
This juxtaposition might suggest that while alignment helps in creating a harmonious discussion, it could sometimes come at the cost of drifting away from the core topic. 
On the other hand, debates that fostered generic chatter or a mix of both alignment and generic chatter not only maintained topical relevance but also delved deeper into the subject matter---even if only slightly. 
This suggests a higher complexity in these debates, possibly leading to a more multifaceted understanding of the subject. 
The nuances in reply patterns across the clusters reiterate the importance of fostering diverse discussion styles in educational settings, recognizing that both consensus and deeper exploration have their unique contributions to the learning experience.

\section*{Acknowledgements}
This research was partially funded by Funda\c{c}\~ao para a Ci\^{e}ncia e a Tecnologia under Project "Debaqi: Factors for promoting dialogue and healthy behaviours in online school communities" with reference DSAIPA/DS/0102/2019 and Instituto Lusófono de Investigação e Desenvolvimento (ILIND) Grant COFAC/ILIND/COPELABS/1/2022.

% ---- OR ------- Uncomment this section to use bibtex
\bibliographystyle{splncs03} % We choose the "plain" reference style
\bibliography{cna_23} % Entries are in the refs.bib file

\end{document}